\pdfoutput=1

\documentclass[11pt]{article}

\usepackage{acl}

\usepackage{times}
\usepackage{latexsym}

\usepackage[T1]{fontenc}

\usepackage[utf8]{inputenc}

\usepackage{microtype}

\usepackage{graphicx}
\usepackage{threeparttable}
\usepackage{amsmath}
\usepackage{amsfonts}
\usepackage{booktabs}
\DeclareMathOperator{\Lagr}{\mathcal{L}}
\DeclareMathOperator*{\argmax}{arg\,max}

\title{Efficient Task-Oriented Dialogue Systems with \\Response Selection as an Auxiliary Task}

\author{
  Radostin Cholakov \\
  High School of Mathematics \\
  Plovdiv, Bulgaria \\
  \texttt{r.cholakov@obecto.com} \\
   \And
  Todor Kolev \\
  Obecto Ltd. \\
  Sofia, Bulgaria \\
  \texttt{tkolev@obecto.com} \\
}

\begin{document}
\maketitle
\begin{abstract}
The adoption of pre-trained language models in task-oriented dialogue systems has resulted in significant enhancements of their text generation abilities. However, these architectures are slow to use because of the large number of trainable parameters and can sometimes fail to generate diverse responses. To address these limitations, we propose two models with auxiliary tasks for response selection - (1) distinguishing distractors from ground truth responses and (2) distinguishing synthetic responses from ground truth labels. They achieve state-of-the-art results on the MultiWOZ 2.1 dataset with combined scores of 107.5 and 108.3 and outperform a baseline with three times more parameters. We publish reproducible code and checkpoints and discuss the effects of applying auxiliary tasks to T5-based architectures.
\end{abstract}


\section{Introduction}

Task-oriented dialogue (TOD) systems are developed to lead conversations with users and assist them with the completion of various tasks. Unlike traditional solutions which rely on natural language understanding, state tracking, language generation, and other modules, end-to-end systems utilize a single network for all required functionality \cite{6407655}. The recent research in the field has concentrated on leveraging language models pre-trained on general-domain corpora \cite{devlin2018bert, radford2019language, raffel2020exploring} to produce more robust architectures fine-tuned specifically for TOD generation. This has bridged the gap between production-ready modularized pipelines and single-network models in terms of accuracy and human-sounding results. However, such architectures are big and computationally expensive; they are also prone to overfitting on the final task and "forgetting" useful capabilities from the pre-training phase \cite{greco2019psycholinguistics, kulhanek2021augpt}. Multiple studies (Section \ref{sec:related}) have demonstrated that learning related auxiliary tasks can improve the generation performance of a model while making it less affected by the overfitting issue. 

In this paper, we study the effects of learning auxiliary response selection tasks together with an architecture based on the T5 \cite{raffel2020exploring} text-to-text transformer. We use MTTOD \cite{lee-2021-improving-end}, trained on the MultiWOZ 2.1 \cite{eric2019multiwoz} dataset, as a baseline and evaluate two main approaches for response selection:

\begin{itemize}
    \item Binary classifier to distinguish between encodings of ground truth responses and encodings of distractor sentences sampled from the dataset.
    \item Binary classifier to tell apart ground truth responses from decoder-generated sequences.
\end{itemize}

Reproducible code and model checkpoints are available at \url{https://github.com/radi-cho/RSTOD}.

\section{Related Work}
\label{sec:related}

TOD sequence-to-sequence models usually generate a belief state based on the dialogue history and then use the belief state (in addition to the previous context) to generate a response \cite{lei2018sequicity}.

\textbf{Pre-trained Language Models} such as BERT \cite{devlin2018bert}, GPT-2 \cite{radford2019language}, and T5 \cite{raffel2020exploring} significantly enhance dialogue systems when they are fine-tuned for sequence tasks. The first study to validate this on GPT-2 is \cite{budzianowski2019hello}. SOLOIST \cite{peng2020soloist}, UBAR \cite{yang2021ubar}, and SimpleTOD \cite{hosseini2020simple} further develop the end-to-end setting of the problem by considering database results and generated responses during training. MinTL \cite{lin2020mintl} provides a minimalistic \textbf{transfer learning} dialogue system with multiple backbones. TOD-BERT \cite{wu-etal-2020-tod} utilizes a contrastive objective function to mimic a response selection task. \cite{yang2022robust} augments data by ignoring nonessential tokens and also adversarially filters “easy” samples to enhance model robustness.

\textbf{Auxiliary Learning} - training additional tasks which improve the performance of the primary text generation task - is increasingly applied in TOD systems. AuGPT \cite{kulhanek2021augpt} demonstrates that response selection tasks are helpful on top of GPT-2. MTTOD \cite{lee-2021-improving-end} has a span selection auxiliary task. GALAXY \cite{he2022galaxy} (with UniLM \cite{dong2019unified} as a backbone) optimizes four objectives, one of which is a selection between ground truth responses and randomly sampled responses. PPTOD \cite{su2021multitask} is also trained for multiple tasks in a plug-and-play fashion \cite{dathathri2019plug}.

\section{Method}
\label{sec:method}

\subsection{Dialogue System Baseline}

\begin{figure*}[h]
    \centering
    \includegraphics[width=12cm]{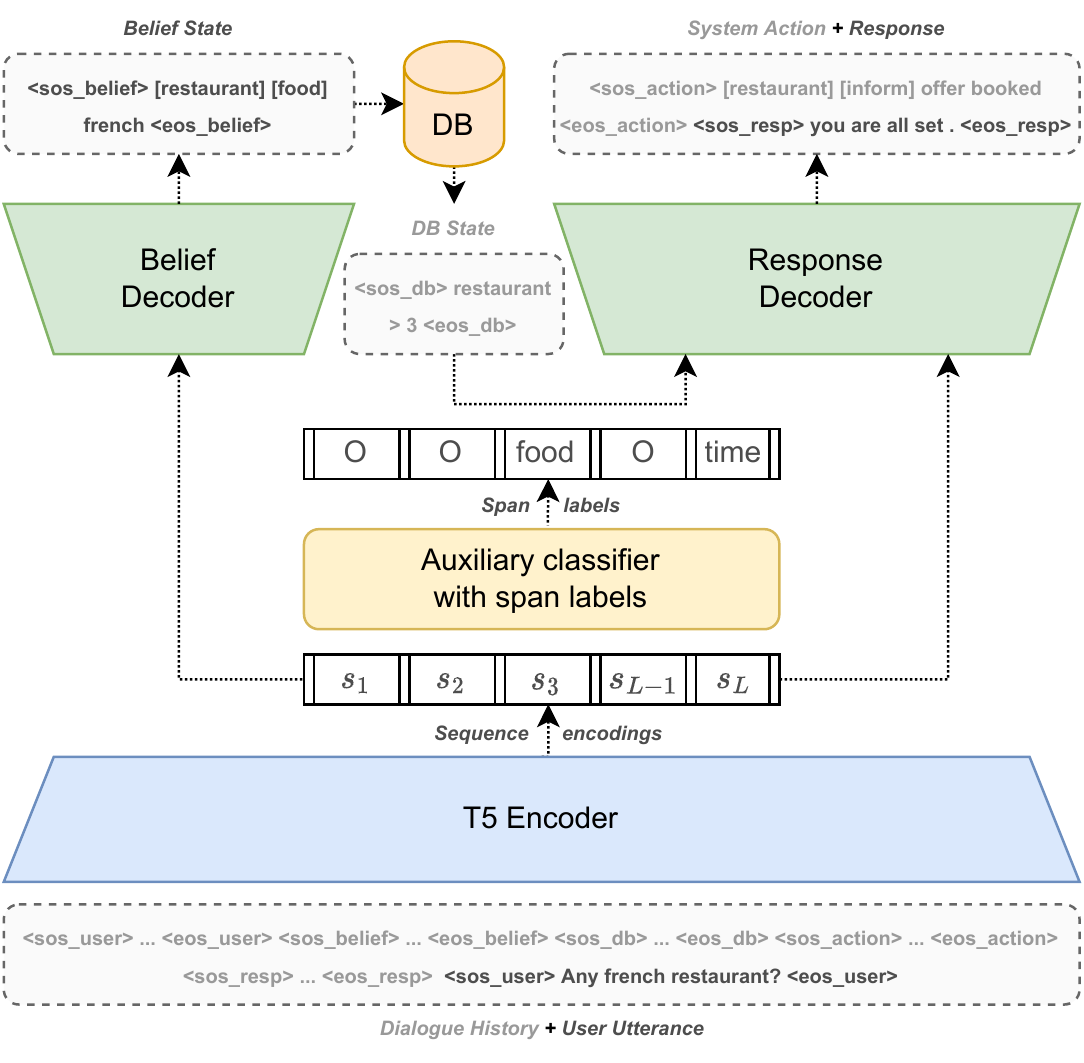}
    \caption{Dialogue generation architecture.}
    \label{fig:MTTOD}
\end{figure*}

As a baseline, we use the end-to-end system setting introduced in \cite{lee-2021-improving-end} (Figure \ref{fig:MTTOD}) with T5 encoder-decoder backbone. The encoder input consists of a dialogue history concatenated with a user utterance. A \textit{belief state} decoder generates a sequence of a domain name, slot names, and slot values. There is an option for querying a domain-specific database based on the belief state to generate a \textit{DB state} which is then used to condition a final \textit{response} decoder. The response output contains a \textit{system action} state - a sequence of a domain name, action types and slots - and a \textit{system response}. Since the decoder works autoregressively\footnote{An autoregressive decoder uses information from previous time steps to generate the value at the current time step.}, response generation is automatically conditioned on the system action.

As shown in figure \ref{fig:MTTOD}, MTTOD utilizes a classifier as an auxiliary task for span matching, inspired by recent dialogue state tracking approaches. Labels for this task are the extractive informable slots defined in \cite{gao2020machine}.

The loss to be jointly minimized is

\begin{equation}
    \Lagr = \alpha \Lagr_{belief} + \beta \Lagr_{resp} + \gamma \Lagr_{span}
\end{equation}

where $\Lagr_{belief}$ and $\Lagr_{resp}$ are negative log-likelihood language modeling losses for the two decoders and $\Lagr_{span}$ is a cross-entropy loss for the span task. For compatibility with \cite{lee-2021-improving-end} the coefficients $\alpha$, $\beta$ and $\gamma$ are set to $1.0$, $1.0$ and $0.5$ respectively. Refer to section \ref{sec:results} for baseline benchmarks.

\subsection{Response Selection as an Auxiliary Task}

\begin{figure*}[h]
    \centering
    \includegraphics[width=11cm]{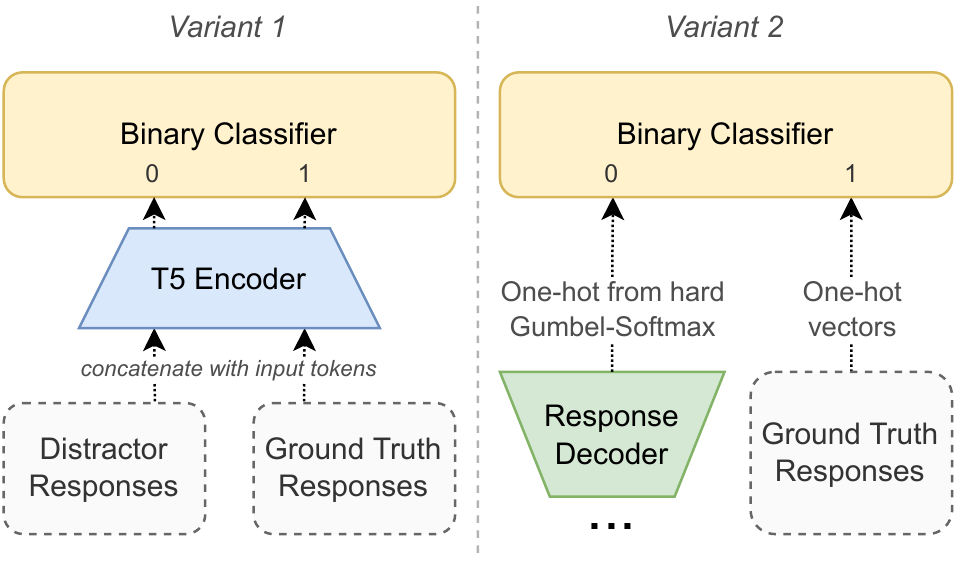}
    \caption{Binary classification response generation tasks.}
    \label{fig:selection}
\end{figure*}

Our study aims to evaluate the effects of using response selection as an additional auxiliary task for the presented T5-based dialogue system. We propose two variants for such a task (Figure \ref{fig:selection}) and modify the full objective to

\begin{equation}
    \Lagr = \alpha \Lagr_{belief} + \beta \Lagr_{resp} + \gamma \Lagr_{span} + \delta \Lagr_{select}
\end{equation}

In our experiments $\delta$ is also set to $0.5$.

\subsubsection{Distinguishing distractor encodings}
\label{sec:distractor}

The first proposal for a response selection task in our system is a binary classifier head - a linear layer or a simple multilayer perceptron - distinguishing randomly sampled \textit{distractor} responses from ground truth responses. During training, the dialogue context $C_t$ at time step $t$ (consisting of history $H_t$ and user utterance $U_t$) is concatenated with both the ground truth labels $T_t$ - forming a sequence $(C_t, T_t)$ - and a distractor response $D_t$ sampled from the dataset - forming a sequence $(C_t, D_t)$. Encodings for both sequences are generated by the already implemented T5 encoder and are then fed to the response selection head. The class label is 0 for $(C_t, D_t)$ and 1 for $(C_t, T_t)$. The binary cross entropy loss to be minimized is defined as

\begin{equation}
    \begin{aligned}
      \mathcal{L}_{select}=-\log p \left(l=1 \mid C_t, T_t\right)\\
      - \log p\left(l=0 \mid C_t, D_t\right)
    \end{aligned}
\end{equation}

\begin{equation}
    \begin{aligned}
    p\left(l=1 \mid C_t, T_t \right) = \\ \operatorname{sigmoid}\left(\phi_a (\phi_E(C_t, T_t))  \right) \in \mathbb{R}^1 \\ \\
    p\left(l=0 \mid C_t, D_t \right) = \\ 1 - \operatorname{sigmoid}\left(\phi_a (\phi_E(C_t, D_t))  \right) \in \mathbb{R}^1
    \end{aligned}
\end{equation}

where $\phi_E$ denotes the encoder and $\phi_a$ - the final classifier.

Optimizing the auxiliary response selection task affects the gradients of the encoder parameters. We empirically prove that this is beneficial for the overall score improvements on multiple metrics.

\subsubsection{Distinguishing generated sequences}

We also propose another independent auxiliary task for response selection inspired by Generative Adversarial Networks \cite{goodfellow2014generative}. Its goal is to distinguish between responses from the transformer $R_t$ and ground truth sequences $T_t$.

The baseline response decoder generates a sequence of token logits $\pi_1$, $\pi_2$, ..., $\pi_k$, where $\pi_i$ is a vector of unnormalized class outputs over a vocabulary with size $v$. To obtain token ids we usually apply

\begin{equation}
    \argmax_{j}{\left[\log \pi_{ij} \right]}, \quad j \in \left[1,v-1\right]
\end{equation}

for every $\pi_i$. However, such a step is not differentiable, and when subsequent layers are optimized, transformer gradients won't be affected, making the auxiliary task useless. One way to overcome the limitation is to re-encode the sequences as previously described in \ref{sec:distractor} and thus backpropagate knowledge to the T5 encoder. Instead, we propose a classifier that works with differentiably sampled token representations and backpropagates knowledge to the whole architecture during training.

We sample vocabulary class probabilities $y_{i1}$, $y_{i2}$, ..., $y_{iv}$ for every token representation $\pi_i$ from a Gumbel-Softmax distribution \cite{jang2016categorical, maddison2016concrete, gumbel1954statistical}:

\begin{equation}
y_{ij} = \frac{\text{exp}((\log(\pi_{ij})+g_j)/\tau)}{\sum_{k=1}^v \text{exp}((\log(\pi_{ik})+g_k)/\tau)}
\end{equation}

where $\tau$ is a temperature, treated as a hyper-parameter, and $g_j$ is a noise sample from a $\text{Gumbel}(0, 1)$ distribution which can be computed by drawing a $u \sim \text{Uniform}(0, 1)$ and applying

\begin{equation}
g= -\log(-\log(u))
\end{equation}

For consistency with ground truth response sequences which are represented with $v$-dimensional one-hot vectors $\hat{y_i}$, we programmatically\footnote{Refer to the \textit{hard} flag in \url{http://pytorch.org/docs/stable/generated/torch.nn.functional.gumbel_softmax}} convert the probabilities $y_i$ to one-hot vectors but compute gradients with their continuous values.

Finally, both $y$ and $\hat{y}$ are fed to the binary classifier $\phi_b$ and the loss to be minimized is computed as

\begin{equation}
    \begin{aligned}
    \mathcal{L}_{select}=-\log p \left(l=1 \mid \hat{y}\right) \\
    -\log p\left(l=0 \mid y\right)
    \end{aligned}
\end{equation}

\begin{equation}
    \begin{aligned}
    p\left(l=1 \mid \hat{y} \right) = \operatorname{sigmoid}\left(\phi_b (\hat{y})  \right) \in \mathbb{R}^1 \\
    p\left(l=0 \mid y \right) = 1 - \operatorname{sigmoid}\left(\phi_b (y)  \right) \in \mathbb{R}^1
    \end{aligned}
\end{equation}

\section{Experiments}

\subsection{Dataset}

In our workflow, we use the large-scale TOD dataset MultiWOZ 2.1 \cite{eric2019multiwoz} for benchmarks and comparisons with baselines.
We follow the preprocessing techniques from \cite{zhang2020task, lee-2021-improving-end} to replace the specific slot values with placeholders. Table \ref{tab:table-data} presents more in-depth details and statistics on the contents of the dataset.

\begin{table*}[h]
 \caption{MultiWOZ dataset statistics}
  \centering
  \begin{threeparttable}
      \begin{tabular}{p{4.7cm}p{3cm}p{3cm}p{3cm}}
        \toprule
        Domain & Train dialogues & Dev dialogues & Test dialogues \\
        \midrule
        Police & 245 & 0 & 0 \\
        Hospital & 287 & 0 & 0 \\
        Attraction & 127 & 11 & 12 \\
        Taxi & 326 & 57 & 52 \\
        Train & 282 & 30 & 33 \\
        Hotel & 513 & 56 & 67 \\
        Restaurant & 1199 & 50 & 62 \\
        Train + Attraction & 883 & 148 & 163 \\
        Hotel + Attraction & 437 & 55 & 50 \\
        Restaurant + Attraction & 396 & 78 & 70 \\
        Restaurant + Train & 875 & 157 & 155 \\
        Restaurant + Hotel & 462 & 59 & 49 \\
        Hotel + Train & 1077 & 149 & 144 \\
        Restaurant + Hotel + Taxi & 454 & 41 & 42 \\
        Restaurant + Attraction + Taxi & 431 & 53 & 59 \\
        Hotel +Attraction + Taxi & 444 & 56 & 42 \\
        \midrule
        Total & 8438 & 1000 & 1000 \\
        \bottomrule
      \end{tabular}
  \end{threeparttable}
  \label{tab:table-data}
\end{table*}

\subsection{Training procedure}

Train/development/test sets are generated with 80\%/10\%/10\% of the samples. We optimize the objectives from section \ref{sec:method} for 15 epochs and report the results from the best performing checkpoint on the development set. In our experiments, we tested different learning rate schedule strategies and found the best results to be achieved with a constant learning rate initialized as $5\times10^{-4}$ with liner warmup for the first 10\% of the samples.

For variant 2 of our architecture, a scheduler is used to linearly decrease the Gumbel-Softmax temperature $\tau$ with each training iteration. The optimal initial value for $\tau$ used to derive the results in Table \ref{tab:table-results} is $4$ and is gradually decreased to $0.8$.

\subsection{Evaluation Metrics}

During inference, the response selection head is not used and the model performs the same way in terms of speed as the T5-small baseline. We compute \textit{BLEU} \cite{papineni-etal-2002-bleu}, \textit{Inform} and \textit{Success} metrics for both architecture variants. \textit{Inform} validates whether system entities are correct and \textit{Success} checks whether relevant information is given for all user inquiries. A combined score is derived as 0.5 × (Inform + Success) + BLEU which is consistent with previous studies.

\section{Results}
\label{sec:results}


\subsection{MultiWOZ Benchmarks}

Table \ref{tab:table-results} compares the calculated benchmarks for the two proposed variants of our auxiliary task. As a baseline, we present the results of MTTOD with \textit{T5 base} backbone, which has more than 360 million trainable parameters. In contrast, our models, which use \textit{T5 small} as a backbone, have 102.2 and 105.5 million parameters but achieve higher overall results with total scores of 107.4 and 108.3, respectively.

\begin{table*}[h]
 \caption{Benchmark results on MultiWOZ 2.1}
  \centering
  \begin{threeparttable}
      \begin{tabular}{llllllll}
        \toprule
        Model & Backbone & Selection task & Parameters & Inform & Success & BLEU & Score \\
        \midrule
        MTTOD\tnote{*} & T5 base & None & 360.9 M & 92.30 & 84.00 & 19.41 & 107.56 \\
        MTTOD\tnote{*} & T5 small & None & 102.2 M & 89.20 & 80.50 & 19.14 & 103.99 \\
        RSTOD (ours) & T5 small & After encoder & 102.2 M & 92.10 & 83.30 & \textbf{19.69} & 107.39 \\
        RSTOD (ours) & T5 small & Differentiable & 105.5 M & \textbf{93.50} & \textbf{84.70} & 19.24 & \textbf{108.34} \\
        \bottomrule
      \end{tabular}
      \begin{tablenotes}\footnotesize
        \item[*] MTTOD benchmarks are reproduced using its public source code. A slight deviation from the results in \cite{lee-2021-improving-end} is caused by a correction in the evaluation scripts as acknowledged on \url{https://github.com/bepoetree/MTTOD}.
      \end{tablenotes}
  \end{threeparttable}
  \label{tab:table-results}
  
\end{table*}


\section{Discussion}

Response selection tasks similar to variant 1 of our architecture have been previously applied in models for chit-chatting and question answering \cite{wolf2019transfertransfo}. For TOD systems such tasks are used in architectures with GPT-2 (AuGPT) and UniLM (GALAXY) backbones resulting in responses with higher text-generation metrics. Our study is the first to provide an in-depth analysis of whether a T5-based model in a task-oriented setting would benefit from selection tasks. The results we present are consistent with related literature since we also observe an increase in generation performance.

Most of the solutions relying on pre-trained language models have big amounts of trainable parameters making them slow to train. In our study, we use a modification of the baseline with T5-small instead of T5-base, reducing the parameters more than 3 times. In variant 1 the shared encoder is responsible for processing more sequences than the baseline - it is slower to train but identical in terms of inference speed and amount of storage space required for its weights. Variant 2 is comparable in terms of train-time and inference-time speed to the baseline but is able to achieve a higher overall score. It employs techniques for overcoming backpropagation issues with the discrete token representations of a generated response sequence\footnote{Usually text is generated by picking the most likely tokens from a probability distribution over the token space. This is not a differentiable operation and prevents gradient computations.}.

\section{Future Work}

Directions for further research on the topic of TOD systems include testing our proposals on bigger backbone models to evaluate their effectiveness against overfitting, experimenting with additional auxiliary tasks for the current baseline, and introducing data augmentations. Also, whether our classifier heads could be used during inference to perform real-time response selection should be explored.

As a long-term development in the field, we consider various possibilities for building production-ready end-to-end dialogue systems by employing reinforcement learning or semi-supervised learning methods. More experimentally, a generative adversarial network for creative text generation could also be tested.

\section{Conclusion}

In this paper, we propose two independent auxiliary tasks for response selection on top of a TOD system transformer baseline. Both tasks demonstrate state-of-the-art results on multiple text generation metrics despite having 3+ times less trainable parameters. The first variant involves a classifier, distinguishing between distractor and ground truth responses, which affects the transformer encoder during training and achieves results consistent with related literature. The second variant applies a novel technique for the TOD problem and involves a classifier, distinguishing between synthetic and ground truth responses. We publish reproducible code implementations of our proposals and present potential directions for future research.

\bibliography{anthology,custom}
\bibliographystyle{acl_natbib}

\end{document}